\documentclass{article}

\usepackage[main, final]{neurips_2025}

\usepackage{amsmath}
\usepackage{caption}
\usepackage{appendix}
\usepackage{tcolorbox} 
\tcbuselibrary{skins} 
\usepackage{enumitem}
\usepackage[linesnumbered, ruled, vlined]{algorithm2e}
\usepackage{subcaption}  
\usepackage[utf8]{inputenc} 
\usepackage[T1]{fontenc}    
\usepackage{hyperref}       
\usepackage{url}            
\usepackage{booktabs}       
\usepackage{amsfonts}       
\usepackage{nicefrac}       
\usepackage{microtype}      
\usepackage{xcolor}         

\title{LLM-Assisted Causal Structure Disambiguation and Factor Extraction for Legal Judgment Prediction}

\author{%
  Yuzhi Liang\thanks{Equal contribution.} \thanks{Corresponding author.} \\
  School of Information Technology \\
  Guangdong University of Foreign Studies \\
  \texttt{yzliang@gdufs.edu.cn} \\
  \And
  Lixiang Ma\footnotemark[1] \\
  School of Information Technology \\
  Guangdong University of Foreign Studies \\
  \texttt{1978320288@qq.com} \\
  \And
  Xinrong Zhu \\
  School of Information Technology \\
  Guangdong University of Foreign Studies \\
  \texttt{20221003196@mail.gdufs.edu.cn} \\
}

\begin{document}

\maketitle

\begin{abstract}
Mainstream methods for Legal Judgment Prediction (LJP) based on Pre-trained Language Models (PLMs) heavily rely on the statistical correlation between case facts and judgment results. This paradigm lacks explicit modeling of legal constituent elements and underlying causal logic, making models prone to learning spurious correlations and suffering from poor robustness. While introducing causal inference can mitigate this issue, existing causal LJP methods face two critical bottlenecks in real-world legal texts: inaccurate legal factor extraction with severe noise, and significant uncertainty in causal structure discovery due to Markov equivalence under sparse features. To address these challenges, we propose an enhanced causal inference framework that integrates Large Language Model (LLM) priors with statistical causal discovery. First, we design a coarse-to-fine hybrid extraction mechanism combining statistical sampling and LLM semantic reasoning to accurately identify and purify standard legal constituent elements. Second, to resolve structural uncertainty, we introduce an LLM-assisted causal structure disambiguation mechanism. By utilizing the LLM as a constrained prior knowledge base, we conduct probabilistic evaluation and pruning on ambiguous causal directions to generate legally compliant candidate causal graphs. Finally, a causal-aware judgment prediction model is constructed by explicitly constraining text attention intensity via the generated causal graphs. Extensive experiments on multiple benchmark datasets, including LEVEN , QA, and CAIL, demonstrate that our proposed method significantly outperforms state-of-the-art baselines in both predictive accuracy and robustness, particularly in distinguishing confusing charges.
\end{abstract}

\section{Introduction}
Legal Judgment Prediction (LJP) aims to automatically predict court rulings based on case fact descriptions, representing one of the most representative core tasks in the field of legal artificial intelligence. In recent years, with the rapid advancement of pre-trained language models (PLMs), fine-tuning approaches based on BERT and its variants have become the mainstream paradigm in LJP research. These methods typically model judgment prediction as a text classification problem, leveraging PLMs' powerful contextual semantic representation capabilities to automatically learn discriminative features related to judicial outcomes from case facts. They have achieved significant performance improvements across multiple authoritative benchmark datasets.

However, from a legal theory perspective, sentencing is not a simple semantic matching or correlation assessment. It is a rigorous reasoning process that strictly follows legal elements and unfolds based on fact-norm causal logic. Existing PLM-based fine-tuning models primarily rely on statistical correlations between case facts and judicial outcomes, lacking explicit modeling of legal elements and their inherent causal structures. Consequently, they are prone to learning spurious correlations unrelated to the adjudicated conclusions. Previous studies have demonstrated that even minor perturbations to punctuation within input factual descriptions—without altering underlying semantics—can significantly alter predictions from state-of-the-art models(\cite{chen2024rethinking}). This phenomenon exposes fundamental limitations in correlation-driven approaches regarding robustness and reasoning credibility.

To address these shortcomings, integrating causal inference into the LJP task has emerged as a key research direction in recent years. Related work seeks to transcend superficial statistical correlations by uncovering more legally meaningful causal mechanisms linking case facts to judicial outcomes. For instance, \cite{liu2021everything} extracted latent causal relationships from case facts and constructed causal graphs to assist in distinguishing easily confused criminal charges; \cite{chen2024rethinking}systematically analyzed bias sources in legal data using Structural Causal Models (SCM) and proposed bias-reduction learning strategies. These studies demonstrate the potential advantages of causal modeling in enhancing model interpretability and robustness.

Despite progress, reliably constructing causal structures within authentic legal text remains a critical bottleneck constraining the practical application of LJP causal methods. Specifically, existing approaches face two primary challenges: (1) Difficulty in accurately identifying legal factors. Keyword extraction methods based on statistical features or unsupervised rules (e.g., YAKE) struggle to distinguish high-frequency narrative components from substantive legal elements, often introducing noise such as names, place names, or template expressions. Methods relying on general open-information extraction tools face severe domain transfer issues, frequently resulting in omissions or errors in extracting key elements within syntactically complex, highly formalized legal texts. (2) Significant uncertainty in causal structures. Traditional causal discovery algorithms rely on observational data. Under the highly sparse conditions of legal text features, statistical independence tests often lack sufficient support. Simultaneously, due to the existence of Markov equivalence, methods like GFCI can only output partial ancestor graphs containing numerous directionally uncertain edges when lacking temporal or intervention information. This forces subsequent models to introduce heuristics or random assumptions, weakening the legal interpretability of causal structures.

To address these challenges, this paper proposes an enhanced inference framework that integrates large language model (LLM) prior knowledge with statistical causal discovery. This approach aims to maintain the rigor of causal inference while leveraging the legal common sense and reasoning capabilities inherent in LLMs. It mitigates noise interference and structural disambiguation issues, achieving a complete closed-loop process from unstructured case facts to causally informed judgment prediction.

Specifically, at the legal factor level, we propose a “coarse-to-fine” hybrid extraction mechanism. This mechanism generates candidate legal factor sets using statistical methods, enhancing coverage of long-tail critical evidence through uniform sampling in word frequency space. Subsequently, by integrating retrieval-enhanced generation and reasoning chain techniques, we guide the LLM to map candidate factors to standard legal elements while systematically filtering non-element noise, thereby constructing a semantically pure, jurisprudentially consistent causal variable space. Secondly, at the causal structure level, we propose an LLM-assisted probabilistic disambiguation and multi-graph sampling strategy. Addressing ambiguous causal edges in the PAG output by GFCI, this paper avoids deterministic or random assumptions. Instead, it employs the LLM as a constrained “soft legal knowledge base.” Under explicit case details and legal provision prompts, the LLM performs prior probability assessments on uncertain causal directions. Combined with judicial logic constraints and fundamental temporal constraints for pruning and sampling, this generates a set of candidate causal graphs consistent with legal principles, thereby preserving and explicitly modeling causal uncertainty. After generating causal graphs, we constrain text attention weights based on graph information to achieve effective integration between causal inference and pre-trained language models.

In summary, the main contributions of this paper can be summarized in three points:
\begin{enumerate}
    \item Proposed a legal factor identification paradigm integrating statistical screening with LLM semantic reasoning, achieving high-precision extraction of legal elements;
   \item Designed an LLM-assisted causal structure disambiguation mechanism to mitigate structural uncertainty caused by Markov equivalence in observational causal discovery;
   \item Constructs a causality-aware judgment prediction model that guides attention mechanisms through explicit causal effects, significantly enhancing accuracy and robustness in distinguishing similar criminal charges.
\end{enumerate}

Experimental results demonstrate that our proposed method achieves significantly higher prediction accuracy than existing mainstream baseline models across multiple legal datasets, including LEVEN, QA, CAIL, LEDGAR, and Overruling.

\section{Related work}
\label{gen_inst}

Early research on legal text classification primarily relied on traditional machine learning methods such as Support Vector Machines (SVM) and Random Forests, combined with feature engineering techniques like TF-IDF and Bag-of-Words for text vectorization. \cite{inbook}employed an SVM classifier for multi-label legal text classification. While emphasizing computational simplicity, this approach captures only superficial information through shallow features, resulting in insufficient generalization for large-scale long documents. \cite{chen2022comparative} compared concept-feature-based Random Forests with deep learning models in legal text classification, finding traditional models retain certain advantages under specific settings. \cite{bifari2024text}explored the application of machine learning techniques (including random forests and XGBoost) in crime category classification based on open legal databases.

With the advancement of deep learning, researchers began applying neural networks to legal text classification to better capture contextual and semantic features. \cite{DBLP:journals/corr/abs-2109-00993} proposed a lightweight LSTM language model pre-training + downstream classification approach for short legal text classification, achieving outstanding performance on short legal text classification tasks. \cite{niklaus2021swiss}introduced the Swiss-Judgment-Prediction benchmark dataset, systematically evaluating multiple BERT models on legal judgment prediction tasks using multilingual judgments from the Swiss Federal Supreme Court. \cite{wang2022d2gclf}developed the D2GCLF framework, which converts legal documents into multi-relational graphs using graph neural networks (GNNs). A graph attention network is then constructed on top of these relational graphs to classify legal documents. \cite{alghazzawi2022efficient}proposed a model combining RFE feature selection with LSTM+CNN for automatic classification of three types of rulings on the U.S. Supreme Court dataset.

In recent years, researchers have explored applying pre-trained language models to legal text classification, such as fine-tuning BERT or RoBERTa. \cite{chalkidis2020legal}introduced LEGAL-BERT, which significantly improved performance on legal tasks by further pre-training BERT on large-scale legal corpora. \cite{limsopatham2021effectively}proposed a hierarchical BERT strategy, segmenting documents and aggregating attention to process lengthy legal texts. \cite{mamakas2022processing} extended LegalBERT by integrating Longformer to expand attention windows and fuse TF-IDF information, enhancing classification performance on long documents.\cite{paul2023pretrainedlanguagemodelslegal}retrained LegalBERT on Indian legal data, creating InlegalBERT—a model tailored to Indian legal vocabulary that improved performance not only in the new domain (Indian texts) but also in the original domains (European and UK texts). \cite{geng2021legaltransformermodelshelp}released LegalRoBERTa, a RoBERTa model further pre-trained on legal corpora. Building upon this, \cite{garcia2024robertalexpt} and \cite{saketos2024largelanguagemodelgreeklegalroberta} introduced RoBERTaLexPT for Portuguese and GreekLegalRoBERTa for Greek, respectively.

With the rapid advancement of large language models (LLMs), researchers have begun transforming legal text classification tasks into generative problems, leveraging prompt engineering and instruction tuning to enhance model generalization. \cite{parizi2023comparative} conducted a comparative study on prompting strategies for legal text classification, experimenting with multiple LLMs and prompting techniques. \cite{niklaus2025lawinstruct} developed the LawInstruct resource, employing instruction-tuned Flan-T5 to handle legal multi-task classification including text classification. This approach achieved a 50\% improvement on LegalBench but may introduce bias for rare tasks. \cite{jung2025courtroom} proposed the Courtroom-LLM framework, which simulates courtroom debates using multiple LLMs to address ambiguous legal text classification, improving decision accuracy on ambiguous tasks. \cite{liu2025sep} designed the SEP-MLDC paradigm, optimizing multi-label legal document classification by extending label semantics and contrastive learning through LLMs, outperforming baselines on EU legislative datasets. \cite{johnson2025improvingaccuracyefficiencylegal} introduced Legal-LLM, a novel approach leveraging LLM instruction execution capabilities through fine-tuning. It reframes multi-label classification as a structured generation task, instructing LLMs to directly output relevant legal categories for given documents. \cite{chen2025debate} introduce a novel legal judgment prediction model based on a debate feedback architecture, integrating an LLM multi-agent debate model with a reliability assessment model.

\section{Methods}
\label{headings}

\begin{figure*}[!htpb]
    \centering
    \includegraphics[width=1.0\textwidth]{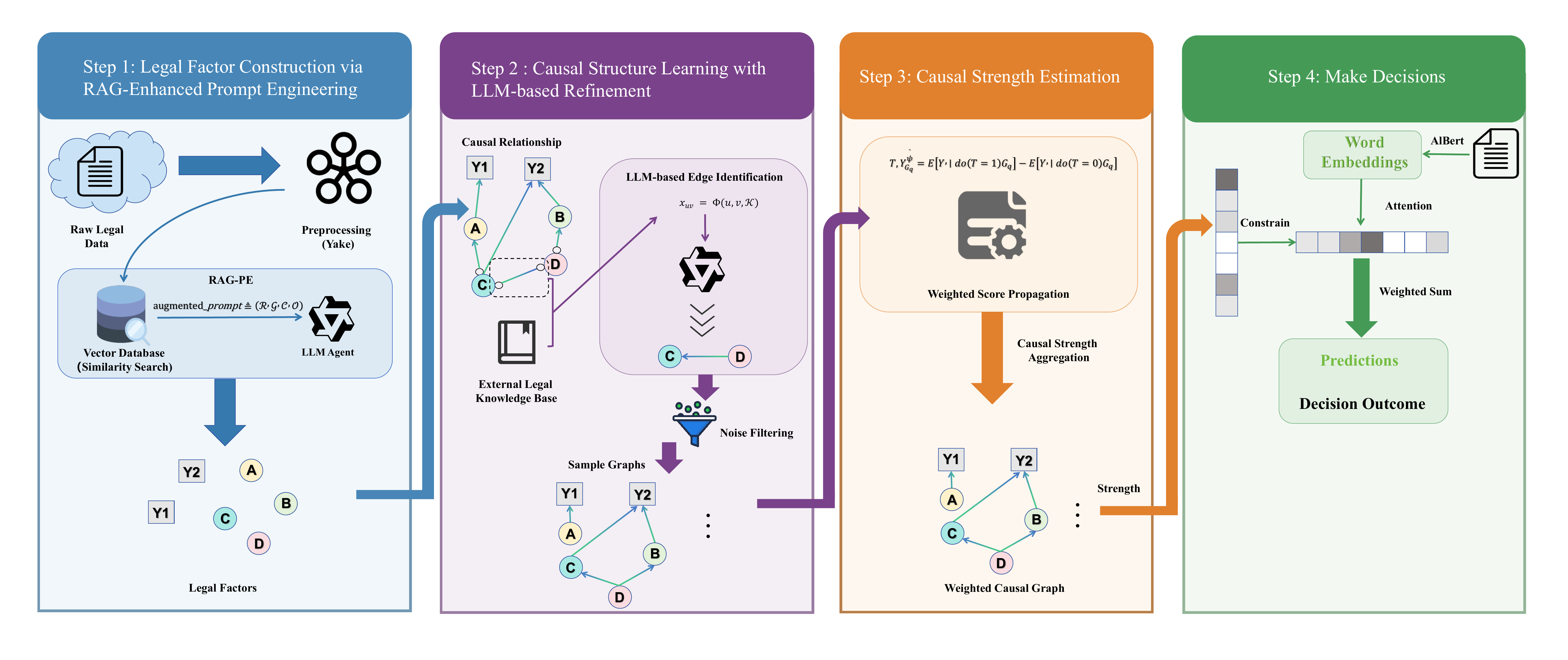}
    \caption{Overall architecture of Ours. For causal diagram nodes, A, B, C, and D represent predicted key nodes, while Y1 and Y2 denote features requiring differentiation.}
    \label{fig:tcanther}
\end{figure*}

To address the issues of causal ambiguity and feature interference in similar charge prediction, this study proposes an enhanced inference framework that integrates Large Language Model (LLM) priors with causal discovery algorithms. As illustrated in Figure \ref{fig:tcanther}, the framework operates in three stages: (1) Legal Factor Discovery and Normalization: Extracting legal factors from unstructured case descriptions via a "coarse-to-fine" paradigm that combines statistical filtering with semantic reasoning; (2) Robust Causal Graph Construction: Leveraging the GFCI algorithm to uncover latent causal structures, while incorporating LLMs as domain knowledge bases to resolve structural uncertainties through probabilistic disambiguation; and (3) Causal Strength Estimation and Decision-making: Quantifying causal effects based on multi-graph sampling and Propensity Score Matching (PSM), followed by an ensemble mechanism for final charge prediction.

\subsection{Hybrid Extraction Mechanism for Legal Factors}
Legal factors serve as the semantic bridge connecting factual circumstances to legal rulings. Addressing the challenge that traditional statistical methods struggle to distinguish core case elements from background noise (e.g., personal names, place names), we designed a hybrid extraction module that combines statistical features with large-model semantic understanding.

\subsubsection{Initial Candidate Set Screening Based on Statistical Features.}
To ensure coverage of long-tail critical evidence, we first employ the unsupervised YAKE algorithm to compute the saliency score $s(w)$ for each lexical item $w$ in the text. YAKE calculates word weights by comprehensively considering multidimensional features including word frequency, position in the text, and contextual relationships. We define the candidate keyword set $\mathcal{K}_{\text{cand}}$ as the subset of words with scores below threshold $\tau$:
\begin{equation}
\mathcal{K}_{\text{cand}} = \{w \in \mathcal{D} \mid s_{\text{YAKE}}(w) \leq \tau\}
\end{equation}
Considering the sparse distribution of features in legal texts, relying solely on frequency truncation introduces noise from high-frequency generic words. Therefore, we define a uniform sampling process on $\mathcal{K}_{\text{cand}}$ to construct the initial set $\mathcal{K}_{\text{init}}$, ensuring unbiasedness in the frequency dimension of the feature space:
\begin{equation}
\mathcal{K}_{\text{init}} \sim \text{Uniform}(\mathcal{K}_{\text{cand}}, N_s)
\end{equation}
where $N_s$ denotes the preset sampling scale.

\subsubsection{Semantic Refinement Based on RAG and Prompt Factor Decomposition}
The initial set $\mathcal{K}_{\text{init}}$ still contains substantial non-legal semantic noise. We leverage the semantic reasoning capabilities of large language models for secondary filtering and introduce Retrieval-Augmented Generation (RAG) to align with standard legal terminology. Specifically, we first introduce an external legal lexicon $\Omega$ as a knowledge base. For each $w \in \mathcal{K}_{\text{init}}$, we obtain its semantic vector using a pre-trained encoder $E(\cdot)$ and retrieve the top-$K$ augmenting contexts $\mathcal{D}_w$ based on cosine similarity:
\begin{equation}
\mathcal{D}_w = \underset{\mathcal{S} \subset \Omega, |\mathcal{S}|=K}{\operatorname{argmax}} \sum_{d \in \mathcal{S}} \cos(E(w), E(d))
\end{equation}

Subsequently, to constrain the LLM's reasoning process, we construct a prompt framework based on element decomposition. This framework is defined as a quadruple $\Phi = (\mathcal{R}, \mathcal{G}, \mathcal{C}, \mathcal{O})$:

\textbf{Role Positioning ($\mathcal{R}$)}: Designates the model as a “criminal law expert,” activating its latent space in the legal domain.

\textbf{Task Goal ($\mathcal{G}$)}: Construct a mapping function $\mathcal{G}: \mathcal{K}_{\text{init}} \times \mathcal{Z} \rightarrow \{0, 1\}$ to determine whether keywords constitute key elements of the target offense group $\mathcal{Z}$.

\textbf{Inference Chain ($\mathcal{C}$)}: The inference chain employs a four-step ordered CoT process, where each subsequent step strictly depends on the preceding results. Specifically, it includes:
\begin{enumerate}
    \item Semantic Parsing: Define the legal interpretation function \(\text{def}: \mathcal{K}_{\text{init}} \times \mathcal{Z} \rightarrow \mathcal{L}\) (where \(\mathcal{L}\) is the legal language space), outputting the unique legal meaning \(\text{def}(k, \mathcal{Z}) = l\) of keyword \(k\) in context \(\mathcal{Z}\), addressing legal term polysemy (e.g., differing legal definitions of “robbery” across contexts).
    \item Element-Relevance Assessment: Based on legal element theory, define the set of constituent elements \(\mathcal{E}_z\) for target offense \(z\). The model analyzes the logical relationship between keywords and elements to determine relevance between keywords and offenses.
    \item Multi-Dimensional Filtering: Introduce a part-of-speech constraint function \(\text{pos}: \mathcal{K}_{\text{initial}} \rightarrow \mathcal{P}\) (where \(\mathcal{P}\) is the part-of-speech space) and an entity type function \(\text{ner}: \mathcal{K}_{\text{initial}} \rightarrow \mathcal{E}\) (where \(\mathcal{E}\) is the entity type space). The filtering rules are:
\begin{equation}
\mathbb{K}_{\text{filtered}} = \{k \in \mathbb{K}_{\text{related}} \mid \text{pos}(k) \in \mathcal{P}_{\text{keep}} \land \text{ner}(k) \notin \mathcal{E}_{\text{exclude}}\}
\end{equation}
where \(\mathcal{P}_{\text{keep}} = \{\text{NOUN}, \text{VERB}, \text{PROPN}\}\) (nouns carrying core legal concepts, verbs expressing legal actions and procedures, proper nouns containing legal terminology), \(\mathcal{E}_{\text{exclude}} = \{\text{PERSON}, \text{GPE}, \text{DATE}\}\) (excludes domain-irrelevant entities like personal names, geopolitical entities, and dates to meet privacy protection needs, reduce overfitting, and avoid temporal interference). 
\item Result Validation and Correction: Introduce a reflection mechanism to simulate the rigor of legal argumentation, validating and correcting screening results to ensure logical soundness.
\end{enumerate}

\textbf{Output Constraint ($\mathcal{O}$)}: Enforce standardized JSON format output for parsing.
See Appendix A for inference chain examples.

We populate the structured prompt template $\Phi(\cdot)$ with $w$, its context $\mathcal{D}_w$, and the target charge group $\mathcal{Z}$. Subsequently, the LLM's reasoning process is modeled as a binary classification task, with its decision function defined as:
\begin{equation}
\mathcal{K}_{\text{refine}} = \left\{ w \in \mathcal{K}_{\text{init}} \;\middle|\; \mathbb{I}\left[ P_{\text{LLM}}\left( y=\text{True} \mid \Phi(w, \mathcal{D}_w, \mathcal{Z}) \right) > \delta \right] = 1 \right\}
\end{equation}
where $P_{\text{LLM}}$ denotes the conditional probability generated by the model, $\mathbb{I}[\cdot]$ is the indicator function, and $\delta$ is the confidence threshold. The inference chain $\mathcal{C}$ enforces the model to follow the logical path “semantic parsing $\rightarrow$ constituent mapping $\rightarrow$ entity denoising,” thereby eliminating non-constituent entities such as proper nouns and place names identified by NER.

\subsubsection{Semantic Clustering and Node Unification}
The refined set $\mathcal{K}_{\text{refine}}$ still contains semantic redundancy (e.g., “steal” and “pilfer”). We encode all words in $\mathcal{K}_{\text{refine}}$ and map them to a numerical space. K-means clustering is then applied to unify nodes, with the number of clusters $K$ dynamically adjusted based on the training set size.

\subsection{Causal Structure Learning Incorporating Expert Knowledge}
After establishing the graph nodes $V$, the core task is to infer the causal structure among nodes. We propose a two-stage strategy combining “algorithmic discovery and knowledge refinement.”

\subsubsection{Initial Structure Discovery}
Given the presence of unobservable confounding factors in legal cases, we employ the GFCI (Greedy Fast Causal Inference) algorithm (\cite{ogarrio2016hybrid}) for structure learning. GFCI outputs a Partial Ancestral Graph (PAG) containing four edge types: (1) deterministic causal edges $u \rightarrow v$; (2) unobserved confounding edges $u \leftrightarrow v$; (3) partially deterministic causal edges $u  \circ \rightarrow v$; (4) fully uncertain edges $u \circ - \circ v$. We define the set of edges to be disambiguated as $\mathcal{E}_{\text{amb}} = \{(u,v) \mid u \circ - \circ v \in \mathcal{E}_{\text{PAG}}\}$.

\subsubsection{LLM-Assisted Probabilistic Disambiguation Mechanism}
To resolve ambiguous relationships in $\mathcal{E}_{\text{amb}}$, we treat the LLM as a soft knowledge base. By integrating legal text as external knowledge, we estimate prior probabilities for causal directions.
For an edge $(u,v)$, we construct a prompt $x_{uv}$ incorporating node semantics and legal context to guide the LLM in evaluating the plausibility of the relation set $\mathcal{R}=\{u \rightarrow v, v \rightarrow u, u \leftrightarrow v\}$, and outputting the optimal relation $e^*_{uv}$.

To leverage the LLM's domain knowledge in resolving edge direction uncertainty, we designed a structured prompt function $\Phi(u, v, \mathcal{K})$. This function maps node pairs $(u, v)$ and relevant legal text $\mathcal{K}$ to instructions $x_{uv}$. Specifically, the prompt consists of four modules:
\begin{itemize}
    \item Role Anchoring: We explicitly instruct the LLM to assume the role of a “legal expert” to ensure causal reasoning strictly adheres to judicial logic and legal interpretation.
    \item Knowledge Injection: To mitigate model hallucination, we embed relevant legal provisions and case facts (denoted as $\mathcal{D}_{context}$) into the prompt, compelling the model to reason based on given evidence.
	\item Decision Constraints: Following the semantic description of input nodes $u$ and $v$, we restrict the model to choose only among three relationship types: forward causality ($u \rightarrow v$), backward causality ($v \rightarrow u$), or no direct causal link ($u \leftrightarrow v$).
    \item Format Control: We enforce the model to output specific tags for parsing convenience.
\end{itemize}
The LLM's output is parsed into the final causal direction $e^*_{uv}$. The complete prompt template is detailed in Appendix B.

Here, $\epsilon \in [0, 1]$ is a hyperparameter. This mechanism allocates most of the probability mass to the LLM's expert judgment while reserving $\epsilon$ of the probability space to explore other possibilities, thereby balancing the utilization of prior knowledge with maintaining statistical robustness.

\subsubsection{Domain Constraint Filtering}
To further enhance graph validity, we introduce two types of constraints to prune the graph structure:
\begin{enumerate}
\item Judicial Logic Constraint: Based on the principle that “facts determine judgments,” edges pointing from charge nodes to fact nodes are prohibited (i.e., $Y \rightarrow X$ is invalid).
\item Temporal Constraints: Based on the axiom “cause precedes effect,” we compute the relative position expectation $\Delta_{pos}(u, v)$ for any two factors $f_u, f_v$ using the training corpus. If $\Delta_{pos}(u, v)$ significantly exceeds the threshold $\alpha$, the direction $f_v \rightarrow f_u$ is prohibited.
\end{enumerate}

\subsection{Causal Graph Sampling}
To address the challenges posed by residual uncertainty in causal relationships within PAGs for subsequent quantification and decision modeling, $Q$ deterministic causal graphs are generated through random sampling. The sampling logic and probability settings for different PAG edge types are as follows: For deterministic causal edges $u \rightarrow v$, retain the edge; for unobserved confounding edges $u \leftrightarrow v$, delete the edge (indicating no direct causal relationship between them); For partially deterministic causal edges $u \circ \rightarrow v$, a binary selection strategy is applied: edges are retained ($u \rightarrow v$) or removed with equal 0.5 probability (no LLM optimization); For fully indeterminate edges $u \circ -\circ v$, a ternary selection strategy is employed, sampling according to the method defined in Section 3.2.2.

The quality of each sampled graph $G_q$ ($q=1,2,...,Q$) is evaluated using the Bayesian Information Criterion (BIC):
\begin{equation}
\text{BIC}(G_q, D) = -2\log L(G_q, D) + k\log N
\end{equation}
where $L(G_q, D)$ is the likelihood value of graph $G_q$ for dataset $D$, $k$ is the number of edges in graph $G_q$, and $N$ is the total number of cases in the dataset. The BIC score serves as the weight for subsequent “causal strength integration,” with better-fitting graphs receiving higher weights.

\subsection{Causal Strength}

\paragraph{Causal Strength Estimation} Since sampled causal graphs inherently contain noise, we optimize the sampled graph by estimating the strength of causal relationships: assigning high weights to edges with strong causal effects, and near-zero weights to edges with no causal relationship or weak effects. We define the Average Treatment Effect (ATE) as the strength of the edge $(T \rightarrow Y)$ in the causal graph $G$, quantified via Propensity Score Matching (PSM):
\begin{equation}
\psi_{G_q}^{T,Y} = \mathbb{E}\left[ Y \mid \text{do}(T=1), G_q \right] - \mathbb{E}\left[ Y \mid \text{do}(T=0), G_q \right]
\end{equation}
where $\text{do}(T=1)$ denotes the intervention operation forcing factor $T$ to be “present.” This value is estimated via Propensity Score Matching (PSM), which pairs samples from the treatment group ($T=1$) and control group ($T=0$) with similar confounding variable distributions to reduce bias.

\paragraph{Aggregating Overall Causal Strength} To synthesize the causal strength across $Q$ graphs, we obtain the overall causal strength between factor nodes and offense nodes by weighting with goodness-of-fit, thereby reducing interference from single-graph noise. For any factor node $T_j$ and charge node $Y_i$, their overall causal strength $\psi_{T_j,Y_i}$ is defined as the weighted sum of corresponding edge causal strengths across sampled graphs, weighted by each graph's fit (quantified using BIC):
\begin{equation}
\psi_{T_j,Y_i} = \sum_{q=1}^Q \omega(G_q) \times \psi_{T_j,Y_i}^{G_q}
\end{equation}
where:
- $\psi_{T_j,Y_i}^{G_q}$ denotes the estimated causal strength of the edge $T_j \rightarrow Y_i$ in the $q$th graph $G_q$. If this edge does not exist in $G_q$, then $\psi_{T_j,Y_i}^{G_q}=0$;
- $\omega(G_q)$ is the weight of $G_q$, calculated as:
  \begin{equation}
  \omega(G_q) = \frac{\exp\left(-\text{BIC}(G_q, D)\right)}{\sum_{q'=1}^Q \exp\left(-\text{BIC}(G_{q'}, D)\right)}
  \end{equation}

\subsection{Decision Making}

After constructing causal graphs and estimating causal strengths, this paper introduces the pre-trained language model ALBERT during the decision phase. The overall causal strength serves as a prior constraint, guiding the model to perform causally aware charge prediction.

For a given case description text, we employ the pre-trained language model ALBERT to encode the text, obtaining token semantic representations (where $d$ denotes the ALBERT dimension):
\[
\boldsymbol{H} = [\boldsymbol{h}_1, \boldsymbol{h}_2, \ldots, \boldsymbol{h}_n], \quad \boldsymbol{h}_i \in \mathbb{R}^d
\]
Subsequently, the attention layer assigns distinct weights $\mathcal\{a_1, a_2, ..., a_n\}$ to each token. Words are then weighted and summed based on these weights to construct the text embedding $\mathbf{v}_{\text{d}}$, ultimately yielding the prediction vector $\mathbf{r}_{\text{cons}}$.
\begin{equation}
a_i = \frac{\exp\left(\boldsymbol{q}^\top \boldsymbol{h}_i\right)}{\sum_{k=1}^n \exp\left(\boldsymbol{q}^\top \boldsymbol{h}_k\right)}
\end{equation}

\begin{equation}
\boldsymbol{v}_d = \sum_{i=1}^n a_i \boldsymbol{h}_i
\end{equation}
where $\boldsymbol{q} \in \mathbb{R}^d$ is a learnable query vector.

In addition to applying cross-entropy loss $\mathcal{L}_{\text{cross}}$ to the prediction vector $\mathbf{r}_{\text{cons}}$, this paper introduces an auxiliary loss $\mathcal{L}_{\text{causal}}$ to guide the attention module using causal strength.
Given the true label $Y_j$, for each token $w$ belonging to legal factor $T_i$, its corresponding causal strength is the overall causal strength $\psi_{T_j, Y_i}$, where $g_i$ represents the normalized result of this causal strength across the entire token sequence.
$\mathcal{L}_{\text{causal}}$ is designed to align attention weights as closely as possible with this normalized strength $g_i$.

\begin{equation}
\mathcal{L} = \mathcal{L}_{cross} + \lambda \mathcal{L}_{causal},
\end{equation}
where
\begin{equation}
\mathcal{L}_{causal} = \sum_{i=1}^n \left(a_i - g_i\right)^2,
\end{equation}
and $\lambda$ is the trade-off coefficient.

\section{Experimental}
In the experimental section, we aim to address the following research questions:
\begin{itemize}
\item \textbf{RQ1:} Does LLM-Knowledge-GCI demonstrate a significant advantage in prediction accuracy compared to existing mainstream legal judgment prediction models?
\item \textbf{RQ2:} Does LLM-Knowledge-GCI demonstrate stronger data efficiency than baseline models in scenarios with scarce training data (low-resource settings)?
\item \textbf{RQ3:} What are the specific contributions of each core component within the framework to the final model's performance?
\item \textbf{RQ4:} Do the causal structures and estimated causal effects uncovered by LLM-Knowledge-GCI remain robust and reliable when confronted with controlled perturbations?
\item \textbf{RQ5:} How does the magnitude of performance improvement brought by the causal reasoning module vary with the richness of legal factors (key evidence) in case texts?
\end{itemize}

\subsection{Experimental Setup}

\subsubsection{Dataset}

The datasets we utilized comprise three Chinese datasets and two English datasets. Statistical information regarding the datasets is presented in Table \ref{tab:dataset_stats}.
\begin{itemize}
\item LEVEN (\cite{yao2022leven}): LEVEN is currently the largest legal event detection dataset and the largest Chinese event detection dataset. It contains 8,116 legal documents annotated with 108 event types, including 64 accusatory events and 44 general events.
\item QA: The QA dataset comprises 200,000 legal consultation questions across 47 categories.
\item CAIL: This dataset comprises five sets of similar criminal charges from the Criminal Law of the People's Republic of China (\cite{criminallaw2017}), which are difficult to distinguish in practical applications (\cite{ouyang1999confusing}). Corresponding factual descriptions were selected from the Chinese AI and Law Challenge (CAIL2018) (\cite{xiao2018cail2018largescalelegaldataset}).
\item LEDGAR (\cite{tuggener2020ledgar}): This dataset originates from filings with the U.S. Securities and Exchange Commission (SEC), all publicly accessible via EDGAR. Each label represents a single subject within the corresponding contractual clause.
\item Overruling (\cite{zheng2021doespretraininghelpassessing}): This dataset, developed in collaboration with Casetext, is sourced from a corpus of judicial opinions. Each label indicates whether a specific legal sentence explicitly expresses an overruling relationship with prior legal precedent.
\end{itemize}

\begin{table}[htbp]
  \centering 
  \small 
  \setlength{\tabcolsep}{4pt}
  
  \captionsetup{
    position=bottom,       
    justification=centering, 
    labelsep=colon,       
    skip=3pt,             
    width=\textwidth       
  }
  
  \begin{tabular}{l l l l}
    \toprule
    Dataset & Task Type & Text Scale (Entries) & Average Length (Characters) \\
    \midrule
    LEVEN & Legal Event Detection & 8,116 & 502 \\
    QA\_corpus & Legal Consultation Question Classification & 203,459 & 38 \\
    CAIL2018 & Criminal Charge Prediction & 30,183 & 671 \\
    LEDGAR & Contract Clause Classification & 80,000 & 113 \\
    Overruling & Precedent Reversal Detection & 2,400 & 27 \\
    \bottomrule
  \end{tabular}
  
  \caption{Dataset Statistics}
  \label{tab:dataset_stats}
\end{table}
\vspace{-0.5cm} 

\paragraph{Baselines}
To fully validate the performance advantages of our model, we selected 12 representative baseline models for comparison:

BiLSTM (\cite{graves2005framewise}) employs a bidirectional recurrent network architecture to encode text sequences. It aggregates bidirectional hidden states to generate text representations for text classification.

BiLSTM+CRF (\cite{huang2015bidirectional}) combines BiLSTM's contextual features with conditional random field (CRF) label transfer modeling, performing sequence classification through structured decoding.
 
BERT (\cite{devlin2019bert}) undergoes deep bidirectional pre-training based on the Transformer encoder. It extracts either the vector at the [CLS] position or the overall sequence representation, then connects to a classification layer to achieve semantic understanding and text classification.

BERT+CRF (\cite{souza2019portuguese}) combines BERT's deep semantic representations with CRF's structured label decoding capability, simultaneously leveraging semantic information and label dependencies to optimize classification results in sequence labeling tasks.

XLM-RoBERTa (\cite{conneau2020unsupervised}) performs cross-lingual pre-training on multilingual corpora, enabling unified text representation and classification through a shared semantic space.

Legal-RoBERTa (\cite{chalkidis2023lexfiles}) further pre-trains the RoBERTa model on legal corpora, enhancing text representations with domain knowledge for legal text classification.

InLegalBERT (\cite{paul2023pretrainedlanguagemodelslegal}) achieves semantic representation and classification of legal texts by further pretraining and fine-tuning LegalBERT on Indian multilingual legal corpora.

NPC (\cite{jiang2023low}) calculates normalized compressed distances between texts using a lossless compressor and performs classification via k-nearest neighbors without parameter training.

LLMEmbed (\cite{liu2024llmembed}) extracts multi-layer semantic embeddings from lightweight LLMs, fuses them into text representations, and employs shallow classifiers for classification.

ProtoLens (\cite{wei2025protolens}) centers on prototype learning, generating semantic prototypes and performing prototype-aware matching to achieve interpretable semantic representations and text classification.

AC-NLG (\cite{wu2020biased}) constructs a counterfactual dual decoder based on causal backdoor adjustment to model different judgment generation distributions. It employs a judgment predictor to select outputs, enabling bias-free court opinion generation.

CASAM (\cite{chen2024rethinking}) employs a causally aware self-attention mechanism. It constructs graph-based interventions to modulate attention weights, distinguishing causal from non-causal information through OIE. Combined with law-specific adversarial attacks, it achieves robust and bias-free legal text prediction.

\subsection{Main Results (RQ1)}
We compare the performance of our model against baseline methods across four datasets in the table. These results are elaborated by addressing the following research questions.

\begin{table*}[!htbp]  
    \centering  
    \captionsetup{justification=centering, labelsep=colon}

    \setlength{\tabcolsep}{8pt}  
    \small  
    \begin{tabular}{lccccc}  
        \toprule  
        Models & LEVEN & QA & CAIL & LEDGAR & Overruling \\
        \midrule  
        BiLSTM & 2.93 & 19.75 & 69.63 & 76.24 & 91.78 \\
        BiLSTM+CRF & 34.91 & 28.24 & 63.90 & 78.38 & 92.51 \\
        BERT & 72.37 & 78.64 & 88.45 & 87.01 & 95.69 \\
        BERT+CRF & 72.94 & 79.17 & 88.81 & 86.35 & 95.88 \\
        Legal-roberta-base & 35.46 & 76.87 & 62.93 & 87.29 & 96.57 \\
        InLegalBERT & 11.78 & 22.97 & 63.54 & 86.74 & 94.12 \\
        xlm-roberta-base & 53.28 & 78.49 & 81.81 & 84.59 & 96.48 \\
        LLMEmbed & 53.63 & 78.89 & 85.23 & 81.47 & 92.78 \\
        npc & 33.53 & 75.66 & 77.02 & 78.16 & 92.18 \\
        ProtoLens & 46.01 & 77.98 & 82.65 & 80.11 & 91.45 \\
        CASAM & 71.08 & 81.37 & 86.19 & 87.21 & 92.75 \\  
        AC-NLG & 74.11 & 83.85 & 88.48 & 82.28 & 89.56 \\  
        zero shot-qwen3:14b & 59.81 & 37.61 & 68.02 & 66.54 & 84.56 \\
        zero shot-llama2:13b & 14.76 & 27.89 & 35.09 & 41.76 & 77.89 \\
        zero shot-gemma3:12b & 44.10 & 37.68 & 66.63 & 56.87 & 83.46 \\
        fine tune-qwen3:1.7b & 21.14 & 73.18 & 80.11 & 82.24 & 91.89 \\
        fine tune-gemma3:1b & 6.81 & 71.84 & 77.97 & 76.28 & 88.76 \\
        ours & 74.26 & 85.72 & 89.31 & 88.31 & 97.05 \\  
        \bottomrule  
    \end{tabular}
    \caption{Comparison of Performance Metrics Across Models}  
    \label{tab:model_performance_50pct}  
\end{table*}

\paragraph{Q1: How does the improved model compare to others in overall performance?}
\ 
\newline 
\indent 
Overall experimental results show that the proposed model achieves optimal or second-best overall performance across varying training set proportions, with particularly stable advantages under medium-to-high resource conditions.

At 10\% or higher training data coverage, traditional sequence models (BiLSTM, BiLSTM+CRF) exhibit low overall performance, with average metrics of 39.54\% and 47.26\% at 50\% training data, respectively. which fails to meet the demands of complex legal text classification tasks. The introduction of pre-trained language models significantly improved overall performance. For instance, BERT+CRF achieved an average metric of 83.02\% with 50\% training data, while XLM-RoBERTa reached 78.31\%.

In contrast, the proposed model achieved superior results under equivalent conditions. With 50\% training data, the improved model achieved an average accuracy of 85.72\% across five datasets, surpassing BERT+CRF and XLM-RoBERTa by 2.70 and 7.41 percentage points respectively. Even with 30\% training data, accuracy reached 84.25\%, demonstrating robust stability and generalization capabilities across varying data scales.

Furthermore, methods based on large model representations or prototype reasoning (e.g., LLMEmbed, ProtoLens, NPC) demonstrate competitive performance in certain scenarios. However, their performance fluctuates significantly across different datasets and remains overall inferior to our model. This further validates the cross-dataset consistency and robustness advantages of our approach.

\paragraph{Q2: What advantages does the improved model offer in few-shot scenarios (1\%, 5\% training sets)?}
\ 
\newline 
\indent 
In few-shot scenarios, the proposed model demonstrates particularly outstanding advantages, indicating its excellent data efficiency and low-resource adaptability.

With only 1\% training data, our model achieves an all-task average accuracy of 61.17\%, significantly outperforming all baselines. It particularly excels on LEVEN: at 1\% data, it achieves 28.89\% accuracy—4.77 percentage points higher than the next-best LLMEmbed (24.12\%) and over ten times better than Bert (1.79\%) and InLegalBERT (1.19\%); It also demonstrates robust performance across C3RD Level 1, QA, and CAIL tasks, achieving accuracy rates exceeding 65\%, with CAIL reaching 71.95\%. When using 5\% training data, the model's average performance further improved to 73.71\%, demonstrating significant advantages over LLMEmbed (65.27\%), NPC (55.07\%), and ProtoLens (51.56\%). This represents a 12.54 percentage point increase compared to the 1\% data scenario, indicating that the performance gains from additional data significantly outperform other models. This outcome demonstrates that our model can effectively mitigate data scarcity by leveraging stable legal factors and causal relationships for reasoning, even with limited annotated samples.

\paragraph{Q3: What are the characteristics of performance improvements across different crime category subtasks?}
\ 
\newline 
\indent 

Based on experimental results across various crime category tasks, the proposed model demonstrates consistent performance advantages in high-similarity crime classification and multi-granularity crime prediction tasks.

For LEVEN, which involves multiple crime categories, baseline models exhibit overall low accuracy. For instance, LLMEmbed achieves 39.58\% accuracy with 10\% training data. In contrast, our model reaches 60.72\% under identical conditions and improves to 74.26\% with 50\% training data—outperforming the second-best baseline Bert+CRF (70.58\%) by 3.68 percentage points, demonstrating stronger discriminative capabilities.

For QA and CAIL, our model maintains high and stable accuracy across varying training data scales, achieving 85.72\% and 89.31\% accuracy respectively with 50\% training data. On QA, our model outperforms all comparison methods, being the only one to exceed 80\% accuracy. The results demonstrate that our model exhibits strong generalization capabilities across different task formats and semantic complexities.

\subsection{Ablation Experiments}

To further validate the effectiveness of key components within the LLM-Knowledge-GCI framework, we designed three sets of ablation experiments examining: (1) the legal factor hybrid extraction module, (2) the knowledge-enhanced module for LLM fuzzy-edge disambiguation, and (3) the causal constraint mechanism module. Results are presented in Table \ref{tab:ablation_experiment}.

\textbf{Legal Factor Hybrid Extraction Module:} This experiment evaluates the effectiveness of combining “traditional keyword extraction (YAKE)” with “LLM semantic enhancement.” Results indicate that the full model outperforms in all task categories: the full model achieves 85.12\% accuracy in the AP\&DD (Abuse of Power \& Dereliction of Duty) group, while the model without LLM enhancement achieved only 82.22\%, a 2.9 percentage point decrease. For the Personal Injury task, the full model reached 90.27\% accuracy, an improvement of 1.2 percentage points. In terms of average performance, the complete model (89.31\%) outperformed the YAKE-only model (87.31\%) by 2 percentage points. This improvement primarily stems from the synergistic advantages of YAKE's statistical feature-based pre-screening and LLM's legal semantics-based refinement. Together, they filter out noise from case descriptions, providing more reliable factor inputs for subsequent causal graph construction.

\textbf{Knowledge Enhancement Module for LLM Ambiguity Resolution:} To validate the necessity of incorporating legal cases and judicial interpretations into prompts for contextual data augmentation, we compared the performance of the full model against a version without the knowledge enhancement module (w/o Knowledge). Results demonstrate that legal knowledge augmentation yields significant gains across multiple crime categories: the full model achieved an average accuracy of 89.31\%, while the version without knowledge augmentation achieved only 87.89\%, a decrease of approximately 1.42 percentage points. Similar trends emerged across multiple crime categories, such as a 2.34 percentage point drop in the EE\&MPF (Embezzlement vs. Misappropriation of Public Funds) task and a 0.76 percentage point decline in the Violent Acquisition task. These findings indicate that legal knowledge provides the model with a clear judicial logic benchmark, preventing potential confusion of legal concepts that might arise from pure textual semantic reasoning.

\textbf{Causal Constraint Mechanism:} Finally, this paper conducted an ablation analysis on the role of the constraint mechanism. Experimental results show that removing the attention module after causal constraints led to a significant decline in overall model performance: on the F\&E (Fraud vs. Extortion) task, accuracy dropped from 93.76\% to 90.88\%, a decrease of 2.88 percentage points; In the EE\&MPF task, the accuracy of the model without the attention mechanism (88.76\%) decreased by 3.76 percentage points compared to the complete model (92.52\%); the average accuracy also dropped from 89.31\% to 86.19\%. These results demonstrate that attention mechanisms with causal constraint injection effectively highlight key information relevant to charge determination during the model's decision-making phase. This provides the necessary alignment foundation for causal constraints to function, thereby enhancing overall performance in distinguishing similar charges.

\subsection{Random Edge Comparison Experiment}

To delve deeper into the mechanism by which the LLM-based \textbf{uncertain edge semantic disambiguation strategy} enhances model performance, this section designs a rigorous controlled variable experiment. This experiment aims to validate the core hypothesis: the performance gains stem from the \textbf{high-quality semantic logic} injected by the LLM, rather than an increase in the causal graph's \textbf{topological density}. To this end, we constructed a randomized baseline strictly aligned with the experimental group in terms of edge count to isolate the influence of quantity on performance.

\paragraph{Control Group Design}
Following the principle of single-variable control, both groups employed the “LLM keyword filtering” method proposed in Section 3.2.1 to construct high-quality legal factor nodes. The sole variable was the selection mechanism for uncertain edges in the PAG graph:
\begin{itemize}
  \item Experimental Group: Employed the “LLM edge selection” strategy. Leveraged LLM legal reasoning capabilities to construct semantically enriched causal graphs.
  \item Control Group: Adopted an \textbf{equal-quantity random sampling} strategy, randomly adding edges to the graph in strict \textbf{quantity parity} with the LLM-selected edges of the experimental group.
\end{itemize}
\paragraph{Dataset:} The CAIL2018 dataset serves as the experimental framework, encompassing five categories of criminal charges: Personal Injury, Violent Acquisition, F\&E, EE\&MPF, and AP\&DD.

\paragraph{Experimental Results and Analysis}
\begin{figure}[!htpb]
    \captionsetup{
       justification=centering
    }
    \centering
    \includegraphics[width=\linewidth]{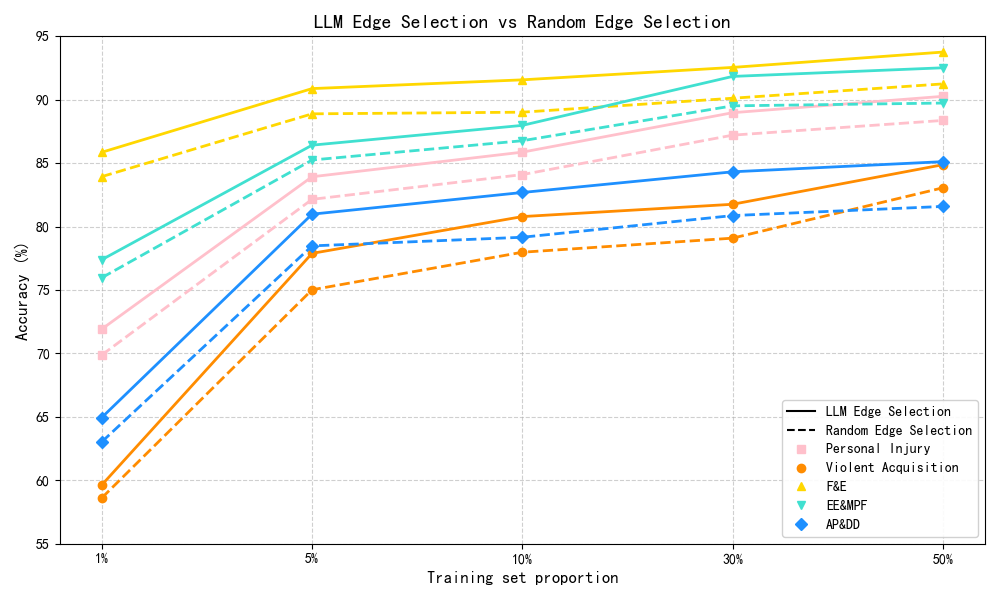}
    \caption{Accuracy Comparison Between LLM Edge Selection and Random Edge Selection Strategies at Different Training Set Ratios}
  \label{fig:random_edge_comparison} 
\end{figure}
Figure \ref{fig:random_edge_comparison} illustrates the accuracy comparison between the LLM edge selection strategy and the random edge selection strategy across different training set proportions. Experimental results demonstrate that models employing the LLM edge selection strategy consistently achieve superior classification performance across different training set proportions and five criminal charge subtasks. Using 50\% training data as an example, the experimental group outperformed the random edge control group in accuracy across all tasks, with the highest reaching 93.76\% (F\&E) and the lowest at 84.88\% (Violent Acquisition).
Even under low-resource scenarios with 1\% and 5\% training data, the experimental group maintained a stable advantage across all tasks. This demonstrates that reasonable causal semantic constraints can provide effective structural priors for models when labeled data is limited. As the training data proportion increases, this advantage persists, indicating that the LLM-guided uncertain edge disambiguation strategy exhibits good adaptability across different data scales.
These results demonstrate that, under consistent causal graph topology scales, the performance gains primarily stem from the high-quality semantic and logical consistency injected by LLMs rather than mere graph structural complexity. This mechanism-level validation confirms the effectiveness of our proposed approach.

\subsection{Causal Graph Quality Analysis}
\begin{figure}[htbp]
    \centering

    \captionsetup[subfigure]{
        justification=centering, 
        labelsep=space, 
        skip=2pt 
    }
    \captionsetup{
       justification=centering
    }
    \begin{subfigure}[b]{0.4\textwidth}
        \centering 
        \includegraphics[width=\textwidth]{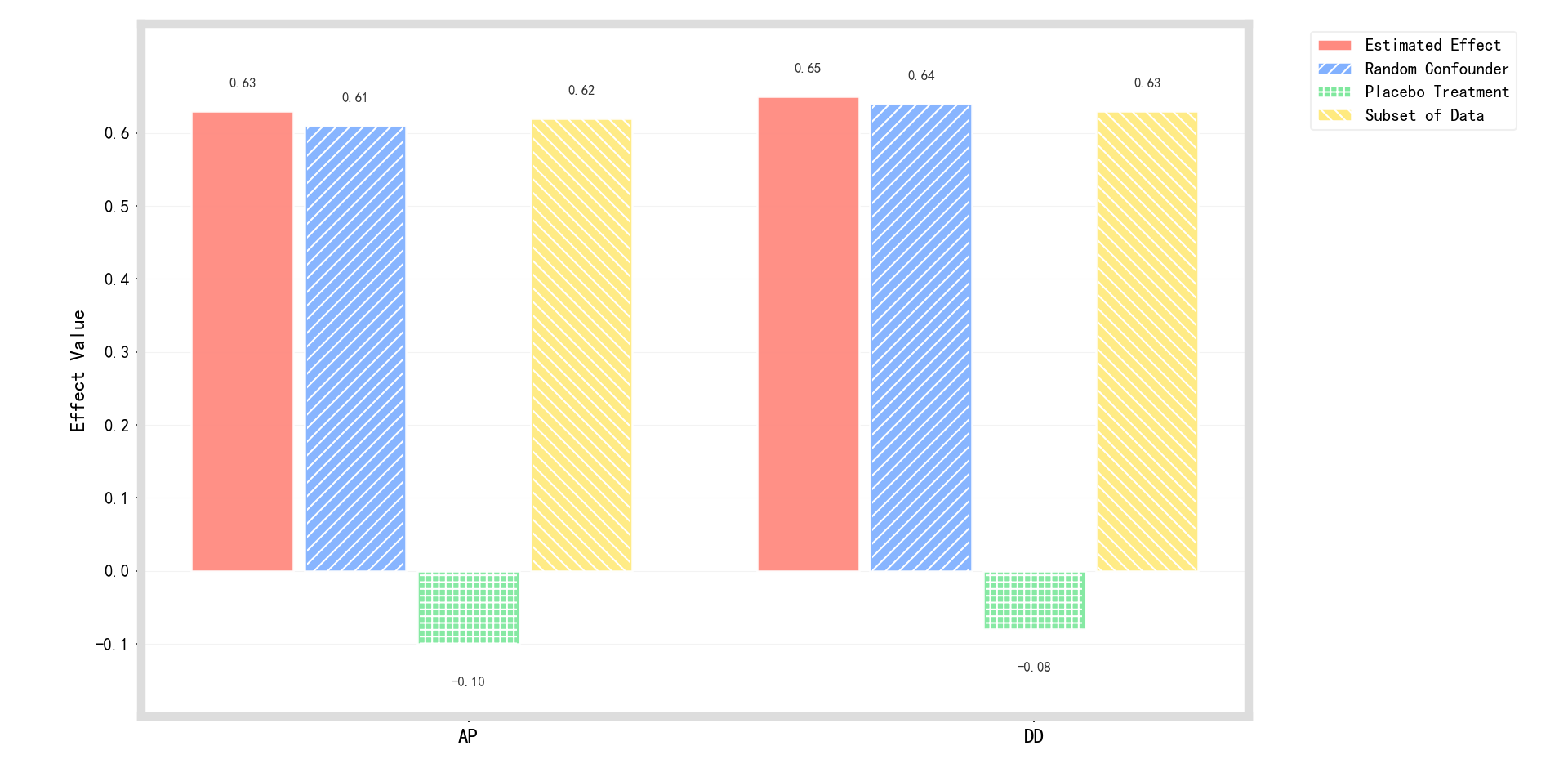}
        \caption{AP\&DD}
    \end{subfigure}
    \hspace{0.01\textwidth} 
    \begin{subfigure}[b]{0.4\textwidth}
        \centering
        \includegraphics[width=\textwidth]{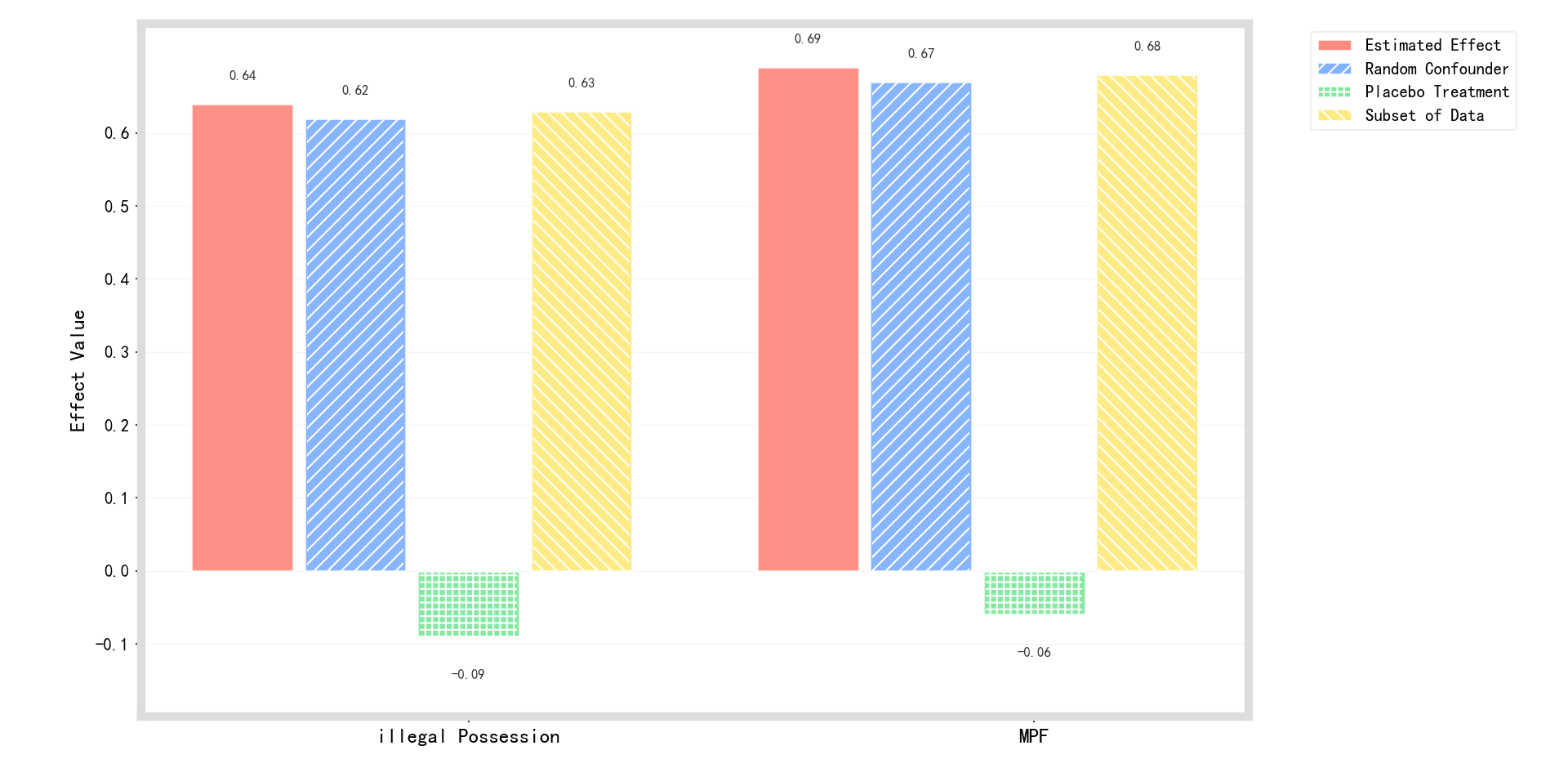}
        \caption{E\&MPF}
    \end{subfigure}

    \vspace{0.5cm} 
    \begin{subfigure}[b]{0.4\textwidth}
        \centering
        \includegraphics[width=\textwidth]{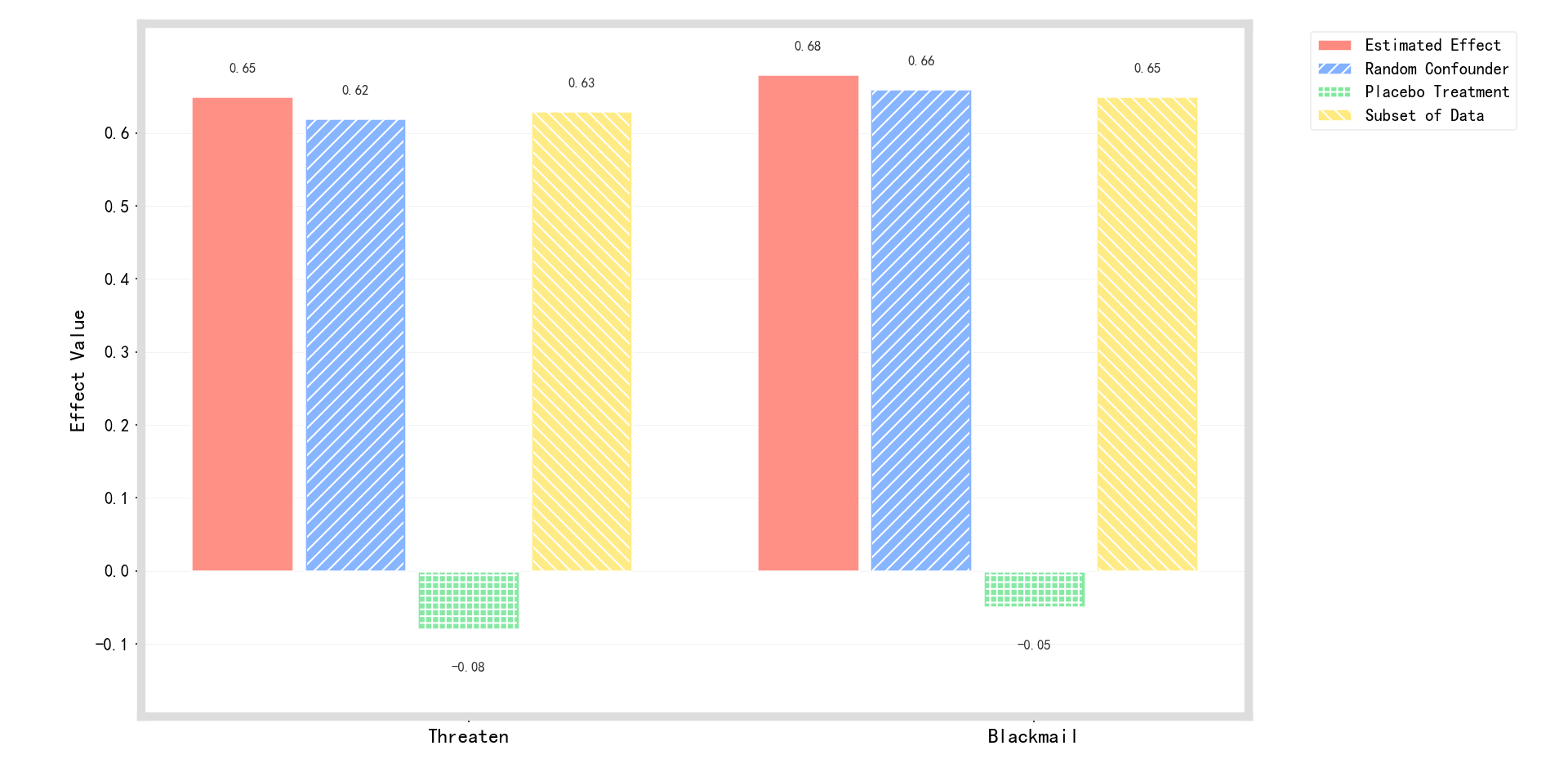}
        \caption{F\&E}
    \end{subfigure}
    \hspace{0.01\textwidth} 
    \begin{subfigure}[b]{0.4\textwidth}
        \centering
        \includegraphics[width=\textwidth]{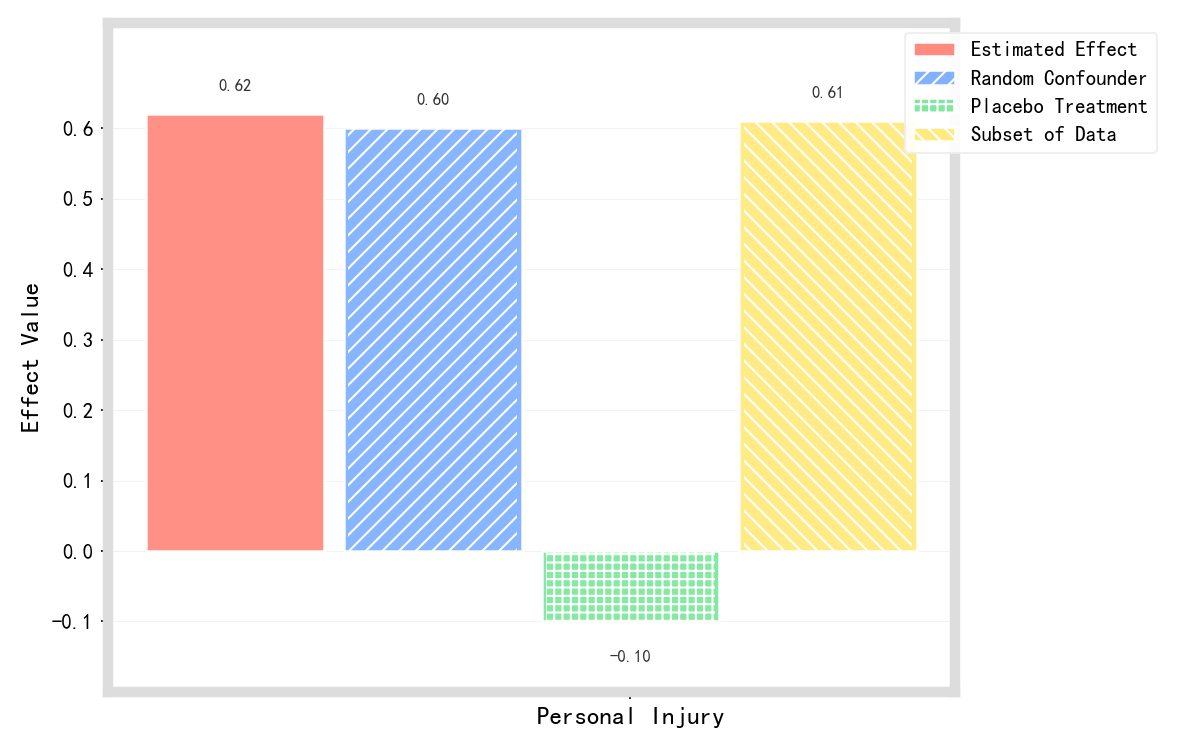}
        \caption{Personal Injury}
    \end{subfigure}
    
    \vspace{0.5cm}
    \begin{subfigure}[b]{0.4\textwidth} %
        \centering
        \includegraphics[width=\textwidth]{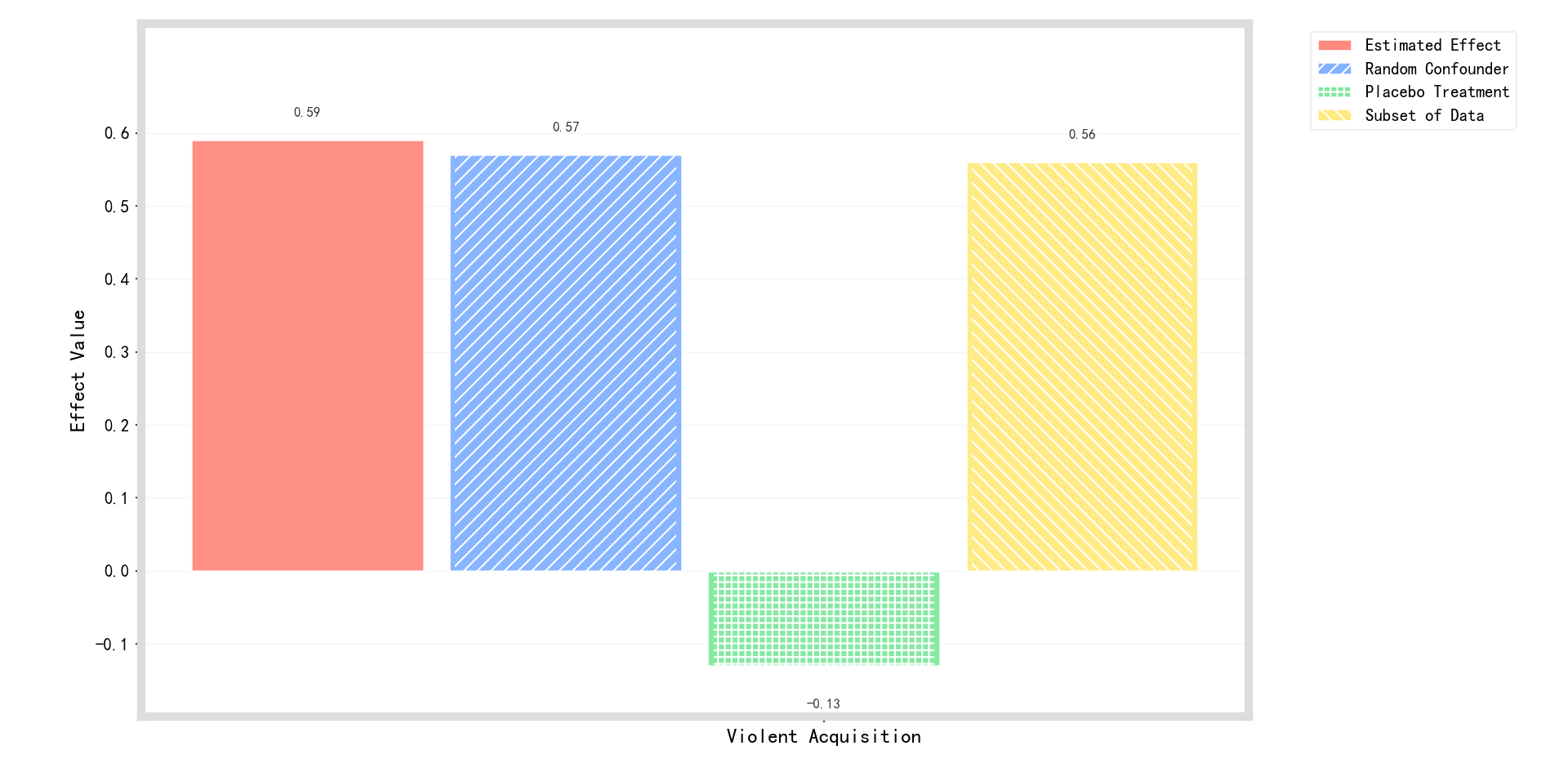}
        \caption{Violent Acquisition}
    \end{subfigure}
    
    \caption{Causal Quality Graph} 
    \label{fig:yinguozhiliang}  
\end{figure}

To validate the robustness of the causal discovery mechanism within this framework, we conducted sensitivity analyses on the generated causal graphs. Specifically, we introduced controlled perturbations to the original causal structure to observe fluctuations in causal effects. Following the evaluation paradigm of Kiciman and Sharm (2018), we designed three refutation strategies: 1) Introduction of random confounding factors: Adding randomly generated confounding variables to the graph, where the original causal effects should theoretically remain constant; 2) Placebo treatment: Replacing the treatment variable with random noise, where the estimated causal strength should approach zero; 3) Subset validation: Re-estimating causal strength on randomly sampled subsets of the data, where the results should not exhibit significant deviation.

Figure \ref{fig:yinguozhiliang} presents experimental results applying these strategies across five crime category tasks. The results indicate that under the random confounding variable and data subset strategies, the causal strength remains largely unchanged; however, after applying the placebo treatment, the causal strength drops to nearly zero. This phenomenon demonstrates that the LLM-Knowledge-GCI framework effectively filters out spurious correlations through knowledge augmentation, enabling the generated causal structure to exhibit superior robustness to perturbations compared to purely statistical methods.

\subsection{Sensitivity Analysis of Relative Model Performance Improvement with Respect to Factor Count}

\begin{figure*}[htbp] 
  \centering 
  \captionsetup{
       justification=centering
    }
  \includegraphics[width=0.9\textwidth]{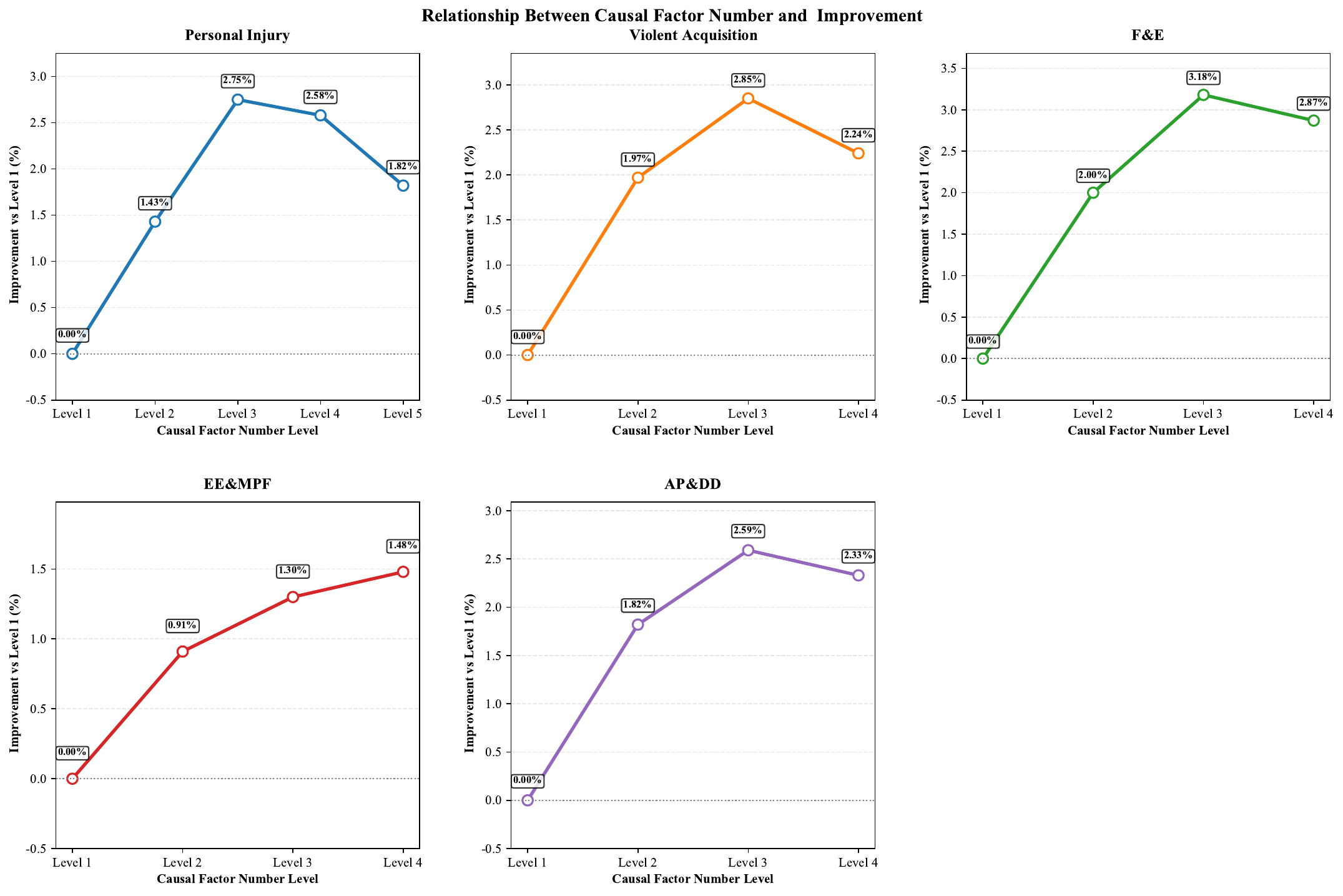}  
  \caption{The Impact of Factor Quality on Outcomes}  
  \label{fig:single_pdf}  
\end{figure*}

To thoroughly investigate the adaptability of the proposed LLM-Knowledge-GCI framework across cases of varying complexity, this section employs the number of causal factors in a case as a quantitative measure of complexity. It systematically analyzes the patterns of relative model performance improvement as the factor count changes, thereby clarifying the framework's performance boundaries under different complexity scenarios.

\subsubsection{Experimental Setup and Grouping Criteria}

Cases were categorized into distinct levels based on factor count to ensure the analysis focuses on complexity's impact on model efficacy. Grouping criteria were determined by: factor distribution patterns within the 50\% training set, the range of legal requirement variations across charge categories, and statistical validity principles:
- Personal Injury: Given case volume, complexity, and requirement variation dimensions, a five-level classification was adopted to precisely segment scenarios from minimal to extreme complexity; Violent Acquisition, F\&E, EE\&MPF, and AP\&DD feature concentrated differences with focused factor counts. A four-level classification maintains distinction while avoiding statistical redundancy from excessive categorization, with each level spanned by 5 factors.

\paragraph{1. Grouping Criteria:}

\begin{itemize}
    \item Five-level classification: Level 1 = [0,4], Level 2 = [5,9], Level 3 = [10,14], Level 4 = [15,19], Level 5 = $\geq20$
    \item Four-level division: Level 1 = [0,4], Level 2 = [5,9], Level 3 = [10,14], Level 4 = $\geq15$
\end{itemize}

\paragraph{2. Metrics:}
Relative Accuracy Improvement:
\[
\Delta Acc_{\text{rel}} = Acc_{\text{Level}i} - Acc_{\text{Level1}}
\]
where $Acc_{\text{Level1}}$ denotes the model accuracy at the lowest factor quantity level (Level 1), and $Acc_{\text{Level}i}$ denotes the model accuracy at level $i$.

\subsubsection {Experimental Results and Analysis}
\ 
\newline 
\indent 
As shown in Figure \ref{fig:single_pdf}, the relative performance improvement ($\Delta Acc_{\text{rel}}$) across datasets exhibits a nonlinear relationship with the number of factors, following these patterns:

\paragraph{1. Complexity Gains and Diminishing Marginal Returns}
\ 
\newline 
\indent 
Except for EE\&MPF, the remaining four datasets exhibit a “rise-then-level-off” trend. Within the Level1 to Level3 range, model performance gradually improves with increasing factor count. For instance, the Violent Acquisition task achieves a 2.85\% relative gain at Level3, indicating that introducing more causal factors helps models better capture critical information structures in cases under moderate complexity scenarios. When the number of factors further increases to Levels 4/5, the performance improvement tends to saturate and even shows a slight decline. This suggests that beyond a certain threshold, excessive factors may introduce redundant information, thereby weakening the model's ability to focus on key discriminative clues, reflecting a trend of diminishing marginal returns.

\paragraph{2. Impact of Offense Specificity on Model Sensitivity}
\ 
\newline 
\indent 
\begin{itemize}
\item \textbf{High Sensitivity (F\&E):} These offenses involve complex subjective intent and multi-step behavioral chains. Increasing the number of factors effectively supplements critical logical nodes, enhancing the model's discriminative capability, resulting in the largest relative improvement.
\item \textbf{Low Sensitivity and Special Cases (EE\&MPF):} This dataset shows a relatively small overall improvement (peak of 1.48\%), but exhibits a stable, monotonically increasing trend as the number of factors increases. This phenomenon may be related to the complexity of legal definitions and semantic expressions for these offenses. While increasing the number of factors provides supplementary information, the overall performance gain remains limited due to the inherent difficulty of the task.
\end{itemize}

\bibliographystyle{unsrtnat}   
\bibliography{ref}             
\section*{Appendix}

\paragraph{A Yake+LLM Prompt Template}
\ 
\newline 
\indent 
\begin{tcolorbox}[
    colback=gray!5, 
    colframe=black!40, 
    boxrule=1pt, 
    arc=2pt, 
    title={System Prompt Quadruple \(\text{system\_prompt} = (\mathcal{R}, \mathcal{G}, \mathcal{C}, \mathcal{O})\)},
    fonttitle=\small\bfseries, 
    center title 
]

\textbf{Role Definition (\(\mathcal{R}\)):} You are a professional legal expert with specialized knowledge in criminal law, required to rigorously screen terms related to target offenses based on legal logic.

\textbf{Task Objective (\(\mathcal{G}\)):} From a given initial keyword set, screen legal terms relevant to the target offense group \(\mathcal{Z} = \{\text{intentional injury}, \text{intentional homicide}\}\).

\textbf{Inference Chain (\(\mathcal{C}\)):}
\begin{enumerate}[leftmargin=2em, itemsep=3pt] 
    \item Analyze the specific meaning of each keyword within the criminal law context (e.g., distinguish between “intentional” and ‘negligent’ harm for “injury”);
    \item Determine whether keywords relate to essential elements of criminal charges (e.g., “intentional subjective state,” “harmful act,” “harmful consequence”);
    \item Retain nouns, verbs, and proper nouns; exclude personal names, place names, and dates;
    \item Verify if the filtered results cover core elements; rescreen if any are omitted.
\end{enumerate}

\textbf{Output Constraints (\(\mathcal{O}\)):} Output only a JSON-formatted list of relevant terms, with a minimum of 15 entries. Example format: [“Term1”, “Term2”, ...].
\end{tcolorbox}

\paragraph{B Legal Node Causality Determination Prompt Template}
\ 
\newline 
\indent 
\begin{tcolorbox}[
    title={Legal Node Causality Determination Prompt Template},
    fonttitle=\small\bfseries,  
    center title,
    left=12pt,
    right=12pt,
    top=12pt,
    bottom=12pt,
    colframe=black!40,
    colback=white
]

\begin{tcolorbox}[colframe=gray!40, colback=gray!5]
\textbf{System Prompt} \\
As a legal causation determination expert, you must integrate legal provisions, precedents, and judicial interpretations. Based on the legal probability of each relationship type, select the relationship type with the highest probability to ensure results align with judicial logic.
\end{tcolorbox}

\vspace{6pt}  

\begin{tcolorbox}[colframe=gray!40, colback=gray!5]
\textbf{Task Prompt} \\
1. Data Augmentation and Legal Basis:\\
   To enhance judgment accuracy, please first review the legal text data and consult relevant legal resources; [Legal Data Text]

\vspace{4pt}
2. Node Semantics:
\begin{itemize}[leftmargin=1.5em, itemsep=1pt, label={-}]
    \item Node $u$ denotes ‘forgery’
    \item Node $v$ denotes ‘acquisition’
\end{itemize}

\vspace{4pt}
3. Core Task: Select the relationship type with the highest probability of causality between nodes $u$ and $v$ from the following:
\begin{itemize}[leftmargin=1.5em, itemsep=1pt, label={-}]
    \item ‘$u \rightarrow v$’: Direct causal relationship from $u$ to $v$
    \item ‘$v \rightarrow u$’: Direct causal relationship from $v$ to $u$
    \item ‘$u \leftrightarrow v$’: No direct causal relationship
\end{itemize}

\vspace{4pt}
4. Output Requirements:
\begin{itemize}[leftmargin=1.5em, itemsep=1pt, label={-}]
    \item Output only the selected option without any additional text (e.g., explanations, punctuation)
    \item Must be enclosed within <START> and </START>
    \item Example: <START>$u \rightarrow v$</START>
\end{itemize}
\end{tcolorbox}

\vspace{6pt} 

\begin{tcolorbox}[colframe=gray!40, colback=gray!5]
\textbf{LLM Output} \\
<START>u→v</START>
\end{tcolorbox}

\end{tcolorbox}

\paragraph{C Small-Sample Example Set (\(\mathcal{E}_{\text{demo}}\))}
\ 
\newline 
\indent 
\begin{tcolorbox}[
    colback=gray!5, 
    colframe=black!40, 
    boxrule=1pt, 
    arc=2pt, 
    title={Input-Output Mapping Example Strategy},
    fonttitle=\small\bfseries, 
    center title 
]

Input: [“Candidate Word 1”, “Candidate Word 2”, “Candidate Word 3”] \\
Output: [“Candidate Word 1”, “Candidate Word 2”]

\end{tcolorbox}

\paragraph{D Causal Diagram Generation Algorithm}
\ 
\newline 
\indent 
\begin{algorithm}[h]
\SetAlgoLined
\SetKwInOut{Input}{Input}
\SetKwInOut{Output}{Output}
\SetCommentSty{textrm} 
\footnotesize 

\Input{Set of case fact descriptions $D = \{d_1, d_2, \dots, d_N\}$ and set of similar charges $C = \{c_1, c_2, \dots, c_M\}$}
\Output{A set of sampled causal graphs $\{G_1, G_2, \dots, G_Q\}$}
\caption{Causal Graph Generation}
\BlankLine

$K_{raw} \leftarrow \text{YAKE}(D); K_{sample} \leftarrow \text{Sample}(K_{raw}, 150)$\;
$K_{clean} \leftarrow \text{LLM\_RAG}(K_{sample}, \text{THUOCL}, \text{repeat}=3)$\;
$\text{FactorSet} \leftarrow \text{Optimize\_silhouette}(\text{K-means}(K_{clean}, \text{dynamic\_k}))$\;

\BlankLine

$PAG \leftarrow \text{GFCI}(\text{FactorSet}, D)$\;

\BlankLine

\For{$e \in PAG.\text{edges}$}{
    \If{$\text{type}(e) = "\circ-\circ"$}{
        $P_{llm}[e] \leftarrow \text{Calibrate}_{\epsilon}(\text{LLM}(e, \text{legal\_clauses} + \text{law\_cases}))$\;
    }
}

\BlankLine

$PAG \leftarrow \text{Filter}(PAG, \text{rules}=["\text{No Charge}\to\text{Factor}", "\text{Temporal} > \alpha"])$\;

\BlankLine

\For{$q=1$ \KwTo $Q$}{
    $G_q = \emptyset$\;
    \For{$e \in PAG$}{
        \uIf{$\text{each eage } e = "\to"$}{
            $G_q \leftarrow G_q \cup \{e\}$\;
        }
        \uElseIf{$e = "\circ\to"$ \KwSty{and} $\text{random} < 0.5$}{
            $G_q \leftarrow G_q \cup \{e\}$\;
        }
        \ElseIf{$e = "\circ-\circ"$}{
            $r = \text{Sample}(P_{llm}[e])$\;
            \If{$r \in \{"\to", "\leftarrow"\}$}{
                $G_q \leftarrow G_q \cup \{r\}$\;
            }
        }
    }

}
\Return $\{G_1, G_2, \dots, G_Q\}$\;

\end{algorithm}
\end{document}